\documentclass[runningheads]{llncs}
\usepackage[T1]{fontenc}
\usepackage{graphicx,verbatim}
\usepackage{array}
\usepackage{booktabs}
\usepackage{amsfonts}
\usepackage{mathrsfs}
\usepackage{amsmath}
\begin{document}
\title{Time-Lapse Video-Based Embryo Grading via Complementary Spatial-Temporal Pattern Mining}
%

\author{Yong Sun\inst{1} \and
Yipeng Wang\inst{2} \and
Junyu Shi\inst{1} \and
Zhiyuan Zhang\inst{1} \and
Yanmei Xiao\inst{3} \and
Lei Zhu\inst{1} \and
Manxi Jiang\inst{3} \and
Qiang Nie\inst{1}}

\authorrunning{Y. Sun et al.}  

\institute{
The Hong Kong University of Science and Technology (Guangzhou) \\ Shanghai Jiao Tong University \\
Center for Reproductive Medicine, Guangdong Second Provincial General Hospital
}

\maketitle              
\begin{abstract}
Artificial intelligence has recently shown promise in automated embryo selection for In-Vitro Fertilization (IVF). However, current approaches either address partial embryo evaluation lacking holistic quality assessment or target clinical outcomes inevitably confounded by extra-embryonic factors, both limiting clinical utility. To bridge this gap, we propose a new task called \textbf{Video-Based Embryo Grading} - the first paradigm that directly utilizes full-length time-lapse monitoring (TLM) videos to predict embryologists' overall quality assessments. To support this task, we curate a real-world clinical dataset comprising over 2,500 TLM videos, each annotated with a grading label indicating the overall quality of embryos. Grounded in clinical decision-making principles, we propose a Complementary Spatial-Temporal Pattern Mining (\textbf{CoSTeM}) framework that conceptually replicates embryologists' evaluation process. The CoSTeM comprises two branches: (1) a morphological branch using a Mixture of Cross-Attentive Experts layer and a Temporal Selection Block to select discriminative local structural features, and (2) a morphokinetic branch employing a Temporal Transformer to model global developmental trajectories, synergistically integrating static and dynamic determinants for grading embryos. Extensive experimental results demonstrate the superiority of our design. This work provides a valuable methodological framework for AI-assisted embryo selection. The dataset and source code will be publicly available upon acceptance.


\end{abstract}
\section{Introduction}
Infertility has become a significant global issue \cite{inhorn2015infertility}, with in-vitro fertilization (IVF) as the primary treatment. The key to IVF success is selecting high-potential embryos, traditionally done through embryologist manual assessments on microscopy images of in-vitro fertilized embryos. However, the evaluation process remains inherently time-consuming and subject to a certain level of inter-observer variability, prompting the development of AI-assisted automated embryo selection systems \cite{DL1,DL2,DL3,DL4,vid_clinical1}. Existing approaches \cite{img_day5_evaluation1,img_viability1,DL2,DL1} primarily focused on the automated analysis of isolated microscopic images. However, in-vitro fertilized embryos require multiple days of dynamic developmental progression before transfer. Consequently, image-centric methodologies overlook holistic developmental information and fundamentally disregard morphokinetic features, such as cell division timing, which are essential for comprehensive evaluations of embryo quality.

The advent of Time-Lapse Monitoring (TLM) technology has enabled the continuous recording of embryonic development process, providing essential morphological and morphokinetic cues for embryo selection \cite{TLM}. A few studies \cite{vid_day51,vid_day52,vid_formation1,vid_stage_classification1,vid_clinical2} have explored video-based methods for automating specific aspects of embryo selection. Some \cite{vid_day51,vid_day52} focused on blastocyst quality assessment using day 5–6 video clips, while others aimed to predict blastocyst formation from early-stage TLM videos \cite{vid_formation1,vid_formation2}. Additionally, research \cite{vid_stage_classification1,vid_stage_classification2} utilized TLM videos for developmental stage classification by modeling temporal correspondences. Although these methods have demonstrated great potential, they nevertheless fail to provide a prediction of the overall embryonic quality necessary for clinical decision-making regarding transplantation. More recently, Kim \textit{et al.} sought to bypass expert evaluation by directly predicting clinical outcomes for automated embryo selection, integrating TLM videos with electronic health records (\textbf{EHR}) based on ViViT \cite{vivit}. However, clinical outcomes are often confounded by a variety of external factors (\textit{e.g.}, uterine environment, maternal age), which increases modeling complexity and leads to limited prediction accuracy. Currently, a common clinical practice for selecting embryos among clinicians is to first evaluate the embryonic development process by holistically analyzing TLM videos, synthesizing morphological and morphokinetic features according to \cite{istanbul}. Embryos with high developmental potential are then prioritized for transfer. Considering this, a viable alternative for automated embryo selection is directly predicting embryologists' holistic evaluation results based on TLM videos, which is clinically translational and remains unexplored.

To close this gap, we propose a new task named \textbf{Video-Based Embryo Grading} - a data-driven paradigm that automatically categorizes in-vitro fertilized embryos into three quality grades (poor/fair/good) by holistically analyzing TLM videos. This paradigm takes TLM videos as input and systematically models embryologists' comprehensive evaluation criteria, generating quality assessments aligned with clinical standards. It holds the potential to seamlessly integrate with time-lapse monitoring systems, enabling an automated and comprehensive embryo selection process. This integration will greatly reduce the burden on clinicians by minimizing the need for manual video analysis.

To support this task, we curate a dataset of 2,596 TLM videos from real-world clinical IVF practices, ensuring that it well reflects the diversity and complexity of everyday clinical scenarios. Each video is paired with a grading label indicating the overall quality of the corresponding embryo. A notable characteristic of this dataset is the variation in video recording lengths, which stems from the natural inconsistencies in embryonic development rates and various clinical operational circumstances. This presents AI systems with two challenges: adaptive temporal modeling to detect critical morphokinetic events without manual annotations, and robust spatiotemporal reasoning to identify fragmented morphological features across various developmental stages for comprehensive quality assessment.

As a powerful baseline, we introduce a clinically motivated framework termed CoSTeM that jointly models complementary morphological and morphokinetic patterns for grading embryos. Built upon CLIP pretrained visual encoder \cite{clip}, we first employ a Bypass Adaption Network to modulate and integrate multiple intermediate features for each frame. These features are decomposed and processed through two specialized streams: 1) A \textbf{morphological branch} that progressively identifies and consolidates discriminative morphological patterns through Mixture of Cross-Attentive Experts Layer and Temporal Selection Block, and 2) a \textbf{morphokinetic branch} that explores embryo developmental trajectories via Temporal Transformer. Extensive comparisons with state-of-the-art embryo selection systems and general video action recognition models demonstrate the superiority of our approach. In summary, our contributions are three-fold:

\begin{enumerate}
    \item We introduce a new task termed video-based embryo grading that predicts embryologists' holistic embryo evaluations directly from TLM videos, accompanied by a curated benchmark dataset to catalyze automated embryo selection research in reproductive medicine.
    \item We propose a clinically inspired dual-branch framework CoSTeM that synergistically models complementary morphological and morphokinetic patterns for automatically grading in-vitro fertilized embryos.
    \item We conduct comprehensive experiments demonstrating the superiority and the rationality of our designs, establishing a foundational benchmark for future research in the field of AI-assisted embryo selection.
\end{enumerate}

\begin{figure}[t]
\centering
\includegraphics[width=0.90\textwidth]{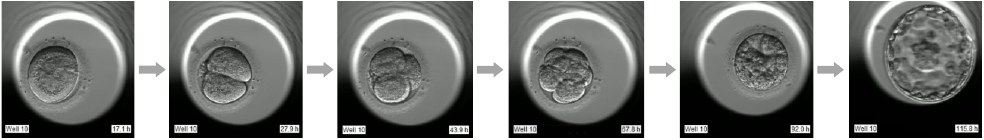}
\caption{Representative frames extracted at key developmental stages of a embryo from "good quality" category} \label{fig:dataset}
\end{figure}

\section{Dataset}
Our dataset contains 2,596 TLM videos from $\ast \ast \ast$ Hospital’s IVF program, collected by the EmbryoScope system. While each video spans over 110 hours post-fertilization, the recording periods exhibit natural temporal fluctuations. 

Each video in our dataset has been meticulously categorized into three classes (poor/fair/good) by embryologists through a systematic, multi-stage evaluation process, incorporating both morphological and morphokinetic criteria. Embryos classified into the high-quality category exhibit excellent features at all stages and possess the best developmental potential. The dataset is partitioned into training and validation subsets in a 7:3 ratio using stratified sampling to preserve class balance. Fig. \ref{fig:dataset} shows some representative frames extracted from key developmental milestones of a good-quality embryo.

\section{Methods}

\begin{figure}[t]
\centering
\includegraphics[width=0.95\textwidth]{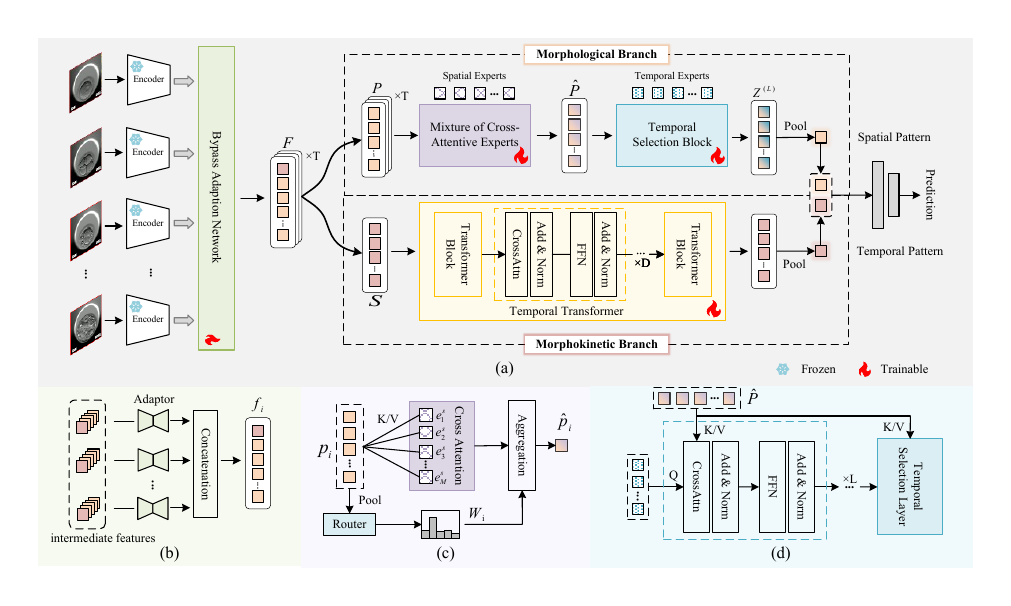}
\caption{Architecture of the proposed CoSTeM framework. (a) Schematic overview of our framework. (b) Bypass Adaption Network for feature modulation and integration. (c) Mixture of Cross-Attentive Experts (MCAE) layer for adaptive spatial feature selection. (d) Temporal Selection Block (TSB) for local frame selection.} \label{fig:network}
\end{figure}

\subsection{Overall Framework}

The fundamental concept of our CoSTeM framework is to emulate the evaluation principle of embryologists for prediction by concurrently mining both static structural characteristics and dynamic developmental patterns.

Given a TLM sequence $V = \{x_t\}^{T}_{t=1}$ with $T$ frames, each frame $x_i$ is partitioned into $N$ spatial tokens and prepended with a $[cls]$ token. These tokens are sent to a pretrained image encoder to extract hierarchical features. To reduce domain gap and construct holistic representations for frames from various developmental stages, a Bypass Adaptation Network, as shown in Fig. \ref{fig:network} (b), modulates intermediate features from layers $\mathcal{L} = \{3, 6, 9, 12\}$ through MLP adaptors \cite{adaptor}, resulting in frame feature sequence $\mathbf{F} = \{f_t\}_{t=1}^T,\; f_t \in \mathbb{R}^{(N+1)\times C}$, where $C = 768$ is the feature dimension. Subsequently, $\mathbf{F}$ is decomposed into two complementary streams: a patch token sequence $\mathbf{P} = \{p_t\}_{t=1}^T, \; p_t \in \mathbb{R}^{N \times C}$ encoding local spatial details for the \textbf{Morphological Branch}, and a class token sequence $\mathbf{S} \in \mathbb{R}^{T \times C}$ capturing global semantics for the \textbf{Morphokinetic Branch}.

The morphological branch aims to selectively capture critical morphological cues (\textit{i.e.}, spatial pattern) from $\mathbf{P}$. Specifically, a Mixture of Cross-Attentive Experts (MCAE) layer dynamically selects discriminative spatial features per frame using $M$ spatial experts, followed by a Temporal Selection Block to identify critical local frames using $K$ temporal experts. 

The morphokinetic branch aims to capture global developmental trajectories (\textit{i.e.}, temporal pattern) using $\mathbf{S}$. In particular, a Temporal Transformer consisting of $D = 2$ cascaded Transformer Block is employed as in ViViT \cite{vivit} to perform global temporal modeling through self-attention. 

Ultimately, the spatial pattern and temporal pattern are concatenated together along the channel dimension and sent to an  MLP classifier for prediction.

\subsection{Mixture of Cross-Attentive Experts}

During embryonic development, various stages exhibit distinct discriminative features. To adaptively focus on stage-specific morphological features, we propose the Mixture of Cross-Attentive Experts (MCAE) layer, inspired by the Mixture of Experts \cite{moe}. As in Fig. \ref{fig:network} (c), a set of spatial experts $\mathbf{E^s} = \{e^{s}_1, ..., e^{s}_M\} \in \mathbb{R}^{M \times C}$ is defined, each conceptually representing a feature selector of a particular embryonic developmental stage. For the $i$-th frame's patch tokens $
p_i \in \mathbb{R}^{N \times C}$, each expert $e^{s}_j$ acts as a query to attend to $p_i$ for feature selection:
\begin{equation}
\mathbf{H}_{i}^{[j]}=\operatorname{CrossAttn}\left(e_{j}^{s}, p_{i}\right) \in \mathbb{R}^{C}, \quad \forall j \in\{1, \ldots, M\}
\end{equation}
Concurrently, a MLP router calculates belonging weights $\mathbf{W}_i \in \mathbb{R}^M$ based on the frame's global averaged context. The $\mathbf{W}_{i}^{[j]}$ approximates the probability of $i$-th frame belongs to $j$-th developmental stage. Afterwards, the final condensed feature for $i$-th frame is a weighted combination:
\begin{equation}
\hat{p}_i=\sum_{j=1}^M \mathbf{W}_{i}^{[j]}\cdot \mathbf{H}_{i}^{[j]}\in\mathbb{R}^C
\end{equation}

Note that these learnable spatial experts and the cross-attention layer are shared across frames, and feature selection is performed in parallel for each frame. The output of MCAE is spatially selected feature sequence $\mathbf{\hat{P}} = \{ \hat{p}_t\}^{T}_{t=1} \in \mathbb{R}^{T \times C}$. 

\subsection{Temporal Selection Block}
According to clinical guidelines, different frames within the TLM video have varying degrees of decisive impact on the final embryo quality, and the lengthy frame sequence may also contain potential noise. Therefore, after conducting stage-specific feature selection in the spatial dimension, it is necessary to further select critical local frames in the temporal dimension.

Inspired by DETR \cite{detr}, we further define $K$ learnable temporal experts $\mathbf{E^t}=\{ e^{t}_1, ..., e^{t}_K \} \in \mathbb{R}^{K \times C}$ to select key local frames. As depicted in Fig. \ref{fig:network} (d), we consider $\mathbf{Z}^{(0)} = \mathbf{E^{t}}$ as queries and $\mathbf{\hat{P}}$ as key and values, passing them through $L = 2$ Temporal Selection Layer contains cross-attention and feed-forward network,
\begin{equation}
\mathbf{Z}^{(l)}=\operatorname{FFN}\left(\operatorname{CrossAttn}(\mathbf{Z}^{(l-1)},\mathbf{\hat{P}})\right)\quad\mathrm{for}\;\;l=1,2,...,L
\end{equation}
where $\mathbf{Z}^{(L)} \in \mathbb{R}^{K \times C}$ is the final temporally selected features. Through cross-attention, learnable temporal experts can adjust their representations and adaptively select frames that contribute most to the final prediction.

\subsection{Loss Functions}
The training objective combines a standard cross-entropy loss $\mathcal{L}_{cls}$ and a diversity loss $\mathcal{L}_{div}$ that minimizes pairwise cosine similarities between temporal experts $\mathbf{E^t}$ and encourages them to select distinct local frames for richer morphological cues. The total loss function is:
\begin{equation}
\mathcal{L}_{total} = \mathcal{L}_{cls} + \lambda \mathcal{L}_{div}, \;\; 
\mathcal{L}_\mathrm{div}=\sum_{i=1}^K\sum_{j=1}^K\left|\frac{(e_i^t)^\top e_j^t}{\|e_i^t\|\|e_j^t\|}\right|
\end{equation}
Here, $\left |\cdot \right |$ denotes absolute value operation, $\left \| \cdot  \right \|$ denotes $L_2$ norm, and $\lambda$ is the balancing coefficient of the two objectives.

\section{Experiments}

\subsection{Experiment Setup}
\subsubsection{Implementation Details}
CLIP-pretrained ViT-B/16 \cite{clip} is chosen as the feature extractor. The $M$ and $K$ are both set to 8. The $\lambda$ is set to 0.1. The training comprises 50 epochs with a batch size of 8, using the AdamW optimizer and a Cosine learning rate scheduler. The learning rate increases from 1.0e-6 to 2.5e-5 after a 5-epoch warmup. Label smoothing of 0.1 is applied to the cross-entropy loss to mitigate overconfidence. For data processing, we uniformly sample 64 frames at 2-hour intervals between 16 and 142 hours post-fertilization from TLM videos, with zero-padding applied if frames are unavailable at specific time points. Original 500×500 frames are resized to 256×256. During training, we apply random 224×224 cropping and data augmentations including horizontal flip, rotation, and color jitter. At inference, frames are center-cropped to 224×224.

\subsubsection{Evaluation Metrics}
 We trained models on the training set and evaluated their optimal performances on the validation set using Accuracy, Precision, Recall, and F1-score.  The macro-averaged F1-score was prioritized as it balances two critical clinical risks: preventing non-viable transfers (precision-driven) and preserving high-potential embryos (recall-sensitive), while accounting for class imbalance.

\subsection{Comparisons with Advanced Methods}
We benchmarked our CoSTeM against state-of-the-art AI-assisted embryo selection methods \cite{vid_day52,keyframefusion,vid_clinical2} and general video action recognition systems \cite{resnet3D,slowfast,aim} for comparison, implementing all approaches using their official codebases or paper specifications under identical training protocols. As shown in Table \ref{tab:comparisons_64_frames}, our proposed CoSTeM framework achieves the best performance. Compared with previous AI-assisted embryo selection methods that lack tailored designs, our framework demonstrates marked superiority. The introduced embryo grading task necessitates the use of extended frame sequences unlike action recognition which typically analyzes sparse frame sequences, posing challenges to the model's adaptive feature extraction capabilities. Among models for video action recognition, the AIM \cite{aim} performs best, achieving an accuracy on par with ours. However, through selective and complementary pattern mining in the proposed CoSTeM, we surpass AIM in other critical metrics: precision ($+$2.95\%), recall ($+$1.97\%), and F1 score ($+$1.89\%). This underscores the superior efficacy of our clinically inspired framework.

\begin{table}[t]
\caption{The classification performance comparisons between selected methods and our proposed CoSTeM framework.}
\label{tab:comparisons_64_frames}
\centering
\fontsize{8}{10}\selectfont 
\begin{tabular}{p{6cm} >{\centering}p{1.2cm} >{\centering}p{1.2cm} >{\centering}p{1.2cm} >{\centering\arraybackslash}p{1.2cm}}
\toprule
Model & Accuracy & Precision & Recall & F1-score \\
\midrule
\multicolumn{5}{@{}l}{\textit{Methods for AI-assisted Embryo Selection}} \\
Xception-LSTM \cite{vid_day52} & 0.8005 & 0.6726 & 0.6982 & 0.6831 \\
KeyFrameFusion \cite{keyframefusion} & 0.8082 & 0.6880 & 0.7122 & 0.6982 \\
EmbryoViViT \cite{vid_clinical2} & 0.8350 & 0.7197 & 0.7229 & 0.7213 \\
\midrule

\multicolumn{5}{@{}l}{\textit{Methods for Video Action Recognition}} \\
ResNet50-3D \cite{resnet3D} & 0.8350 & 0.7444 & 0.6910 & 0.7039 \\
SlowFast \cite{slowfast} & 0.8120 & 0.6959 & 0.7105 & 0.6994 \\ 
AIM \cite{aim} & 0.8606 & 0.7524 & 0.7448 & 0.7477 \\
\midrule

CoSTeM \textbf{(ours)} & \textbf{0.8606} & \textbf{0.7746} & \textbf{0.7595} & \textbf{0.7618} \\
\bottomrule
\end{tabular}
\end{table}

\begin{table}[t]
\caption{Ablation study results for our key designs.}
\label{tab:ablation}
\centering
\fontsize{8}{10}\selectfont 
\begin{tabular}{p{4.5cm} >{\centering}p{1.2cm} >{\centering}p{1.2cm} >{\centering}p{1.2cm} >{\centering\arraybackslash}p{1.2cm}}
\toprule
\multicolumn{1}{l}{Model} & \multicolumn{1}{c}{Accuracy} & \multicolumn{1}{c}{Precision} & \multicolumn{1}{c}{Recall} & \multicolumn{1}{c}{F1-score} \\ \hline
w.o. Morphological Branch & 0.8082 & 0.6921 & 0.6942 & 0.6862 \\
w.o. Morphokinetic Branch  & 0.8350 & 0.7226 & 0.7475 & 0.7327 \\
w.o. Bypass Adaption Network  & 0.8414 & 0.7283 & 0.7318 & 0.7297 \\
w.o. $\mathcal{L}_{div}$ & 0.8542 & 0.7730& 0.7439 & 0.7487 \\ 
\textbf{CoSTeM} & \textbf{0.8606} & \textbf{0.7746} & \textbf{0.7595} & \textbf{0.7618} \\
\bottomrule
\end{tabular}
\end{table}

\subsection{Ablation Study}

\subsubsection{Effectiveness of Core Designs}
We performed an ablation study by iteratively removing key components. As shown in Table \ref{tab:ablation}, removing any single branch significantly degrades performance. This highlights their complementary roles and validates the framework’s alignment with clinical decision logic. Disabling the Bypass Adaption Network disrupts holistic frame representation, causing performance drops. Removing the $\mathcal{L}_{div}$ obviously reduces performance, showing that temporal expert constraints encourage diverse and complementary features.

\subsubsection{Effectiveness of MCAE}
We evaluated the MCAE layer for selecting stage-specific morphological features using various operations. Table 3 shows that MCAE outperforms others, proving its effectiveness in spatial feature selection. Average pooling treats each patch equally, failing to extract stage-specific features for each frame. Max pooling, on the other hand, is susceptible to noise and can result in information loss. The shared expert approach, which uses only one shared expert across frames, also struggles to capture stage-specific features.

\begin{figure}[t]
\centering
\includegraphics[width=1.0\textwidth]{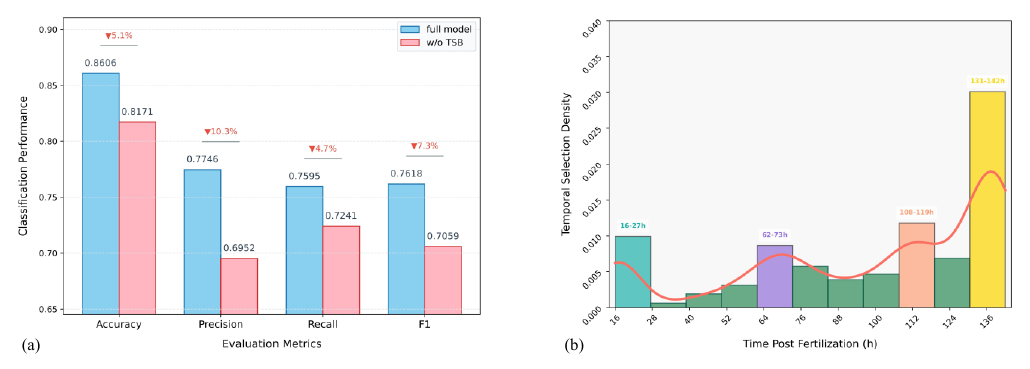}
\caption{Ablation results of TSB. (a) is the performance comparisons of Temporal Selection Block and average pooling. (b) is the temporal density distribution of selected frames by temporal experts in TSB.} \label{fig:TSN}
\end{figure}

\begin{table}[t]
\caption{Comparison results of different frame feature selection methods.}
\label{tab:ablation_spatial_selection}
\centering
\fontsize{8}{10}\selectfont 
\begin{tabular}{p{2.5cm} >{\centering}p{1.2cm} >{\centering}p{1.2cm} >{\centering}p{1.2cm} >{\centering\arraybackslash}p{1.2cm}}
\toprule
\multicolumn{1}{l}{Model} & \multicolumn{1}{c}{Accuracy} & \multicolumn{1}{c}{Precision} & \multicolumn{1}{c}{Recall} & \multicolumn{1}{c}{F1-score} \\ \hline
Average Pooling & 0.8440 & 0.7288 & 0.7411 & 0.7338 \\
Max Pooling & 0.8376 & 0.7205 & 0.7427 & 0.7278 \\
Shared Expert & 0.8453 & 0.7391 & 0.7389 & 0.7385 \\ 
\textbf{MCAE} & \textbf{0.8606} & \textbf{0.7746} & \textbf{0.7595} & \textbf{0.7618} \\
\bottomrule
\end{tabular}
\end{table}

\subsubsection{Effectiveness of TSB}
Fig. \ref{fig:TSN} (a) compares the performance when the Temporal Selection Block (TSB) is replaced with average pooling. Notably, there is a significant decline in all evaluation metrics, highlighting the crucial role of the TSB in accurately assessing embryo quality. Fig. \ref{fig:TSN} (b) shows the density distribution of frames selected by the temporal experts in the first Temporal Selection Layer. Remarkably, these experts mainly focus on key embryonic developmental stages: pronucleus formation stage, cleavage stage, and blastocyst stage, which aligns perfectly with standard embryologist protocols, confirming that our TSB can adaptively capture and select biologically diagnostic features.

\section{Conclusion}
In conclusion, this paper introduces a novel task named Video-Based Embryo Grading, aiming at bridging the divide between current AI-assisted embryo selection paradigms and clinical practices, thereby enabling end-to-end, trustworthy automated embryo selection. To underpin this task, we further curate a dataset that aligns with real-world clinical scenarios. Additionally, we propose CoSTeM, a robust baseline model inspired by clinical decision-making principles. Experimental results highlight the substantial superiority of our method and underscore its potential to propel advancements in AI-assisted embryo selection.

    



%
%
%
\bibliographystyle{splncs04}
\bibliography{references}
%




\end{document}